\documentclass[letterpaper, 10 pt, conference]{ieeeconf}  %

\IEEEoverridecommandlockouts                              %

\usepackage[font=footnotesize,labelfont=normalfont]{caption}
\usepackage{subcaption}
\usepackage{graphicx}
\usepackage{mathtools}
\usepackage{amsmath} %
\usepackage{amssymb}  %

\usepackage{amsthm}
\usepackage{cases}
\usepackage{float}
\newtheorem{theorem}{Theorem}

\theoremstyle{remark}
\newtheorem{remark}{Remark}
\usepackage{siunitx}
\usepackage{cite}
\usepackage{censor}
\usepackage{balance}
\usepackage{hyperref}
\hypersetup{
    colorlinks=true,
    linkcolor=blue,
    filecolor=magenta,      
    urlcolor=cyan,
    pdftitle={Connectivity Maintenance and Recovery for Multi-Robot Motion Planning},
    pdfauthor={Yutong Wang, Lishuo Pan, Yichun Qu, Tengxiang Wang, and Nora Ayanian},
    pdfpagemode=UseNone,
    }

\usepackage[colorinlistoftodos]{todonotes}
\usepackage{booktabs,multirow}
\usepackage{rotating}
\theoremstyle{definition}
\newtheorem{definition}{Definition}[]

\DeclareMathOperator*{\argmin}{arg\,min}

\title{\LARGE \bf
Connectivity Maintenance and Recovery for Multi-Robot Motion Planning
}

\author{%
Yutong Wang, Lishuo Pan, Yichun Qu,
Tengxiang Wang, and
Nora Ayanian%
\thanks{This work was supported by the National Science Foundation under Grant 2330942. Yutong Wang was with the Department of Computer Science at Brown University and is now with the Robotics Institute, Carnegie Mellon University, Pittsburgh, PA 15213 USA. At Brown University, Providence, RI 02912 USA, Lishuo Pan and Nora Ayanian are with the Department of Computer Science, and Yichun Qu and Tengxiang Wang are with the School of Engineering. E-mail:\,\texttt{yutongw3@andrew.cmu.edu}; \texttt{\{lishuo\_pan, yichun\_qu, tengxiang\_wang, nora\_ayanian\}@brown.edu}.}%
}

\begin{document}

\maketitle
\thispagestyle{empty}
\pagestyle{empty}

\begin{abstract}

Connectivity is crucial in many multi-robot applications, yet balancing connectivity maintenance and fleet traversability in obstacle-rich environments remains challenging. Reactive controllers based on control barrier functions can preserve connectivity when it is initially satisfied, but often struggle with deadlocks in cluttered environments. We propose a real-time B\'ezier-based constrained motion planning algorithm, namely MPC--CLF--CBF, that produces trajectories and control inputs concurrently, subject to high-order control barrier function and control Lyapunov function constraints. Our motion planner supports connectivity-aware navigation in cluttered workspaces and recovers connectivity from initially disconnected configurations and after temporary obstacle-induced separation; it also provides analytic continuous-time derivatives, facilitating its application to agile differentially flat systems such as quadrotors. In simulations with $4$--$12$ robots, it maintains $95.8$--$100\%$ graph-connected time at $20\%$ obstacle density, compared with $48.9$--$61.3\%$ for MPC--CBF, with no observed collisions. We further validate the planner in a physical experiment with $8$ Crazyflie nano-quadrotors.

\end{abstract}

\section{INTRODUCTION}

Multi-robot systems such as those in delivery~\cite{scott2017drone}, exploration~\cite{tolstaya2021multirobotcoverageexplorationusi}, and search-and-rescue~\cite{drew2021multi}
rely on communication and sensing within a limited range, demanding that teams maintain proximity. %
We consider two robots connected if they are within a range of each other, and
the fleet connected if every robot is reachable from every other robot through such pairwise connections.
However, balancing connectivity maintenance and fleet traversability in a cluttered environment while enabling connectivity recovery remains a challenge in motion-planner design. %

Many existing control barrier function (CBF) based connectivity controllers are limited to preserving connectivity and do not restore connectivity once it is lost~\cite {capelli2020connectivity, ong2023nonsmooth,de2024distributed, bhatia2024decentralized}. 
CBF controllers are reactive, leading to deadlocks in cluttered environments when the need to satisfy both connectivity and collision-avoidance constraints conflicts with their ability to bypass obstacles.
Furthermore, most safety-critical constraints for a mechanical system have a relative degree greater than one. Our approach generalizes constrained motion planning to use high-order control barrier functions (HOCBFs) and high-order control Lyapunov functions (HOCLFs), thereby handling a broader class of systems.

Motivated by the aforementioned challenges and limitations, we design a continuous-time B\'ezier-curve-based motion planner, namely MPC--CLF--CBF, that balances connectivity maintenance and traversability, enables connectivity recovery, and produces derivatives of arbitrary order that benefit the control of agile differentially flat systems, such as quadrotors.
Our motion planner includes connectivity-preserving HOCBF constraints based on algebraic connectivity and HOCLF constraints that facilitate pairwise proximity.
We use a smooth gate function that adjusts the weights of HOCBF and HOCLF slack penalties online, prioritizing connectivity preservation when the team is connected and favoring recovery when connectivity is lost.
Meanwhile, collision avoidance is imposed as a hard HOCBF constraint at the sampled time steps in the planning horizon.
Sampling these constraints provides a computationally tractable approximation of the continuous-time conditions.
The HOCBF and HOCLF constraints are integrated into a receding-horizon, B\'ezier-parameterized trajectory-generation framework that yields smooth, dynamically feasible trajectories with high-order derivatives in real time.
This design allows connectivity to be restored from initially disconnected configurations and after the team temporarily separates while navigating cluttered environments.

\begin{figure}[t]
    \centering
    \includegraphics[width=\linewidth]{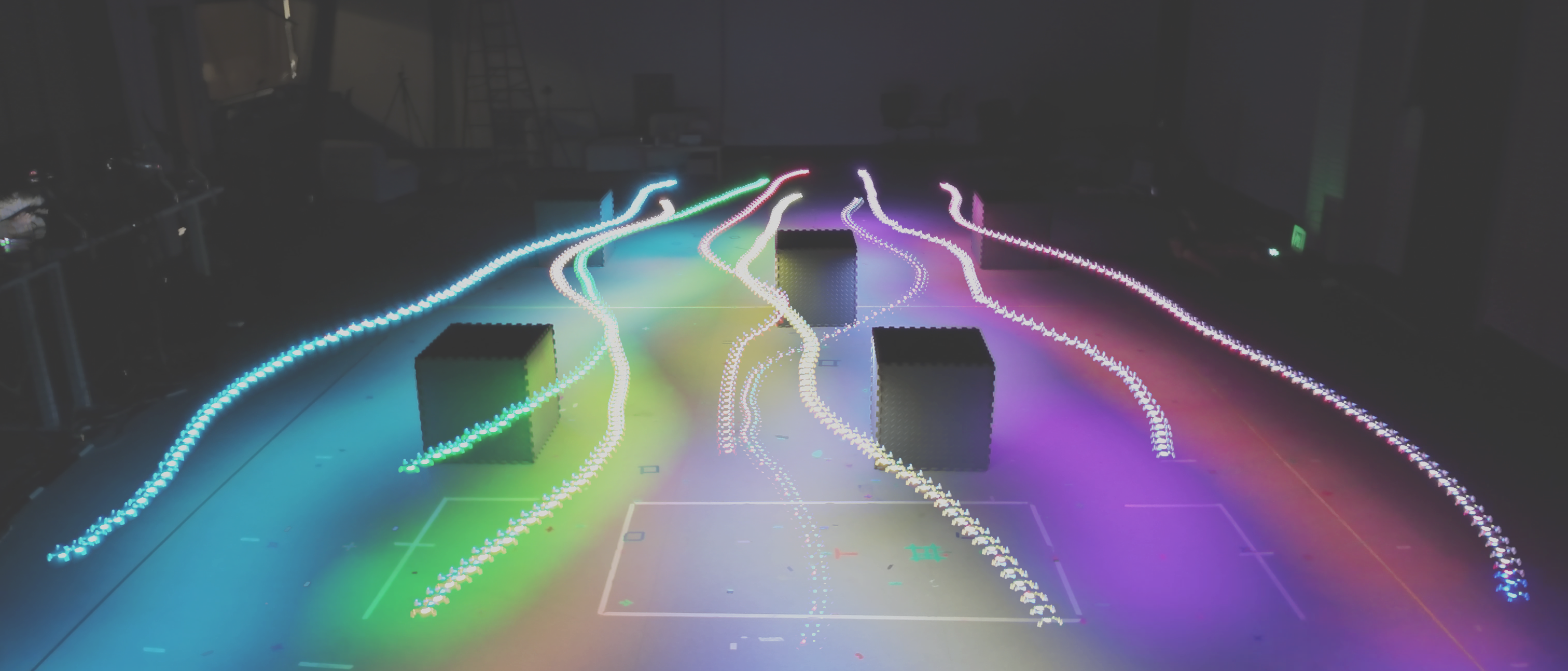}
    \caption{Long exposure of 8 quadrotors navigating a cluttered environment with 5 obstacles.
    }
    \label{fig:demo}
    \vspace{-1.5em}
\end{figure}

The main contributions of this work are %
as follows:
\begin{itemize}
    \item a real-time predictive multi-robot motion planner that balances fleet connectivity and traversability in obstacle-rich environments;
    \item connectivity recovery through HOCLF constraints after temporary disconnections and from initially disconnected configurations, supported by a smooth connectivity-dependent gate that balances maintenance and recovery objectives; and
    \item an MPC--CLF--CBF optimization framework that efficiently integrates HOCBF and HOCLF constraints in a B\'ezier curve-based motion planner. 
\end{itemize}

\section{RELATED WORK}

Connectivity maintenance and recovery in multi-robot systems have been studied through various geometric and optimization-based approaches. 
Early work considers decentralized controllers that emphasize preserving local pairwise links, but these controllers limit reconfiguration and navigation-task performance~\cite{ji2007distributed, dimarogonas2008decentralized}.
Recent work based on algebraic graph theory, particularly the Fiedler eigenvalue, provides connectivity guarantees and flexible fleet configurations via optimization-based control, although it requires global state information~\cite{capelli2020connectivity, ong2023nonsmooth, de2024distributed, bhatia2024decentralized}. These approaches often rely on forward invariance given an initially connected team and lack mechanisms to restore connectivity after disconnection.

Recently, disconnection handling has gained increasing attention, including $k$-connectivity restoration, replanning under environmental changes, and probabilistic search and recollection of lost agents~\cite{ishat2024fast, marchukov2025multi, eshaghi2024restoring}. We address disconnection handling via CLF constraints in a multi-robot motion planning framework. %

CBFs have become a popular model-based tool for safety-critical control due to their guarantees on set invariance without compromising performance~\cite{ames2014Control, ames2019controlbarrierfunctionstheory, xu2019correctness, wang2017safety}. 
In practice, most safety-critical constraints for a mechanical system have relative degrees greater than one. Thus, HOCBFs~\cite{xiao2021high, tan2021high} extend CBFs to a broader class of systems.
Early CBF formulations for connectivity use the Fiedler eigenvalue as a barrier constraint~\cite{capelli2020connectivity, capelli2021decentralized}.
Later works introduce nonsmooth formulations~\cite{ong2023nonsmooth} and distributed CBFs based on local estimates~\cite{de2024distributed, bhatia2024decentralized}. 
However, they only preserve connectivity, employ a simplified first-order kinematic model, and, as reactive CBF-based controllers, are prone to deadlock in obstacle-rich environments.
In multi-robot navigation, the deadlock can be resolved using methods such as multi-agent path finding (MAPF) based trajectory planners~\cite{10801655, pan2025hierarchicaltrajectoryreplanninglarge, tajbakhsh2024conflict}. 
Our approach restores connectivity after loss and integrates CBF constraints into horizon-based trajectory planning, mitigating deadlocks in multi-robot navigation.

Attempts to integrate CBFs into MPC include discrete-time formulations~\cite{zeng2021safety} that impose CBF constraints over a predictive horizon,
as well as a continuous-time formulation using receding-horizon multi-layer controllers~\cite{10644656} and a B\'ezier-curve-based trajectory-planning algorithm constrained by HOCBFs~\cite{11235958}, which concurrently provides continuous-time derivatives.
Relative to reactive CBF controllers~\cite{capelli2021decentralized}, our approach imposes connectivity constraints over a predictive horizon to reduce shortsighted behavior around obstacles. Relative to MPC--CBF~\cite{11235958}, it adds pairwise HOCLF recovery constraints and connectivity-dependent gate weighting.

\section{PRELIMINARIES}
\subsection{High-order Control Barrier Functions and Control Lyapunov Functions}
Consider an affine control system:
\begin{equation}\label{eq:model}
    \dot{\mathbf{x}} = f(\mathbf{x}) + g(\mathbf{x})\mathbf{u}
\end{equation}
where $\mathbf{x} \in \mathbb{R}^p$ is the state of the system, and $\mathbf{u} \in U \subset \mathbb{R}^q$ is the control input, where $U$ is defined as the set of admissible control inputs. A closed set $\mathcal{C}\in \mathbb{R}^{p}$ is \textit{forward invariant} for the closed-loop system $\dot{\mathbf{x}}$ if $\mathbf{x}(0)\in\mathcal{C}\Rightarrow \mathbf{x}(t)\in\mathcal{C}, \forall t$. %
It is assumed that any safe set can be expressed as a zero-superlevel set of a continuously differentiable function $h(\mathbf{x}): \mathbb{R}^p \rightarrow \mathbb{R}$, written as
\begin{align}
    \mathcal{C}=\{\mathbf{x}\in\mathbb{R}^{p} \mid h(\mathbf{x}) \geq 0\}. \label{eq:safe_set}
\end{align}

A control barrier function is a model-based tool used to produce a control input that renders the set~(\ref{eq:safe_set}) \textit{forward invariant} for~(\ref{eq:model}). A typical CBF, however, assumes that $h(\mathbf{x})$ has relative degree one with respect to~(\ref{eq:model}). %

\begin{definition}[Relative degree~\cite{khalil2002nonlinear}]\label{def:relative_degree} A sufficiently smooth function $h$ is said to have \textit{relative degree} $r\in \mathbb{N}$ with respect to~\eqref{eq:model} on the set $\mathbb{R}^{p}$ if 1) $L_{g}L_{f}^{i-1}h(\mathbf{x}) = 0$, for all $1\leq i \leq r-1$, and 2) $L_{g}L_{f}^{r-1}h(\mathbf{x}) \neq 0$ for all $\mathbf{x}\in \mathbb{R}^{p}$.
\end{definition} 

For many mechanical models, the control barrier function $h(\mathbf{x})$ has a relative degree greater than one. To establish \textit{forward invariance} for such models, high-order control barrier functions (HOCBFs) are introduced as follows:

\begin{definition}[HOCBFs~\cite{xiao2021high, tan2021high}] \label{def:HOCBF} Consider a system as in~\eqref{eq:model}. Let $\{\mathcal{C}_{i}\}_{i=1}^{r}$ be a collection of sets of the form $\mathcal{C}_{i}=\{\mathbf{x}\in\mathbb{R}^{p}\mid \psi_{i-1}\geq0\}$ satisfying
\begin{align}
    \psi_{r}(\mathbf{x}) &= \dot{\psi}_{r-1}(\mathbf{x}) + \alpha_{r}(\psi_{r-1}(\mathbf{x}))\\
    \psi_{0}(\mathbf{x}) &= h(\mathbf{x}),
\end{align}
where $\{\alpha_{i}\}_{i=1}^{r}$ is a set of differentiable extended class $\mathcal{K}$ functions. A function $h$ is said to be a HOCBF of order $r$ for~\eqref{eq:model} on an open set $\mathcal{D} \supset \cap_{i=1}^{r}\mathcal{C}_{i}$ if $h$ has relative degree $r$ on some nonempty $\mathcal{R} \subseteq \mathcal{D}$ and there exists a suitable choice of $\{\alpha_{i}\}_{i=1}^{r}$, such that for all $x\in \mathcal{D}$
\begin{align}
    \label{eq:hocbf_const}
    \sup_{\mathbf{u} \in U} [\underbrace{L_f \psi_{r-1}(\mathbf{x}) + L_g\psi_{r-1}(\mathbf{x})\mathbf{u} + \alpha_r(\psi_{r-1}(\mathbf{x}))}_{\psi_{r}(\mathbf{x}, \mathbf{u})}] \geq 0. 
\end{align}
\end{definition}
\begin{theorem}[\!\!\cite{tan2021high}] Let $h$ be a HOCBF for~\eqref{eq:model} on $\mathcal{D}\subset \mathbb{R}^{p}$ as in Def.~\ref{def:HOCBF}. Any locally Lipschitz controller $\mathbf{u} \in U$ that satisfies $\psi_{r}(\mathbf{x}, \mathbf{u})\geq 0$ renders $\cap_{i=1}^{r}\mathcal{C}_{i}$ \textit{forward invariant} for the closed-loop system in~\eqref{eq:model}.
\end{theorem}

Control Lyapunov functions (CLFs) are model-based tools that produce control inputs to stabilize the system at an equilibrium point. Similar to HOCBFs, we extend standard CLFs and define high-order control Lyapunov functions (HOCLFs) as follows:

\begin{definition}[HOCLFs~\cite{xiao2021hoclbf}] Consider a system as in~\eqref{eq:model}. Let $V(\mathbf{x}):\mathbb{R}^{p}\rightarrow \mathbb{R}$ be an $r$-times differentiable function that satisfies the following inequalities for positive constants $c_{1}>0$ and $c_{2}>0$:
\begin{align}
    c_{1}\|\mathbf{x}\|^{2} \leq V(\mathbf{x}) \leq c_{2}\|\mathbf{x}\|^{2}
\end{align}
Define a collection of functions $\{\phi_{i}(\mathbf{x})\}_{i=1}^{r}$ satisfying
\begin{align}
    \phi_{r}(\mathbf{x}) &= \dot{\phi}_{r-1}(\mathbf{x}) + \beta_{r}(\phi_{r-1}(\mathbf{x}))\\
    \phi_{0}(\mathbf{x}) &= V(\mathbf{x}),
\end{align}

where $\{\beta_{i}\}_{i=1}^{r}$ is a set of differentiable extended class $\mathcal{K}$ functions. A function $V(\mathbf{x})$ is said to be a HOCLF of order $r$ for~(\ref{eq:model}), such that for all $\mathbf{x}$,
\begin{align}
    \label{eq:prel-hoclf}
    \inf_{\mathbf{u} \in U} [\underbrace{L_f \phi_{r-1}(\mathbf{x}) + L_g\phi_{r-1}(\mathbf{x})\mathbf{u} + \beta_r(\phi_{r-1}(\mathbf{x}))}_{\phi_{r}(\mathbf{x}, \mathbf{u})}] \leq 0. 
\end{align}

\end{definition}

\subsection{Algebraic Connectivity}
\label{sec:prel-conn}
We represent the multi-robot system as an undirected graph $\mathcal{G} = (\mathcal{V}, \mathcal{E})$, where vertex $v_{i} \in \mathcal{V}$ corresponds to a robot, and edge $e_{i,j} \in \mathcal{E}$ exists if and only if the Euclidean distance between the two robots does not exceed the maximum connectivity distance $R$, i.e., $d_{i, j} = \|\mathbf{p}_i - \mathbf{p}_j\| \leq R$, where $\mathbf{p}_i$ represents the position of robot $i$. As in~\cite{yang2010decentralized}, we can quantify the global connectivity of the system using \emph{algebraic connectivity}, or the Fiedler value~\cite{fiedler1973algebraic}, $\lambda_2$. First, we define the adjacency matrix of the system $\mathcal{A} \in \mathbb{R}^{N \times N}$, where

\begin{equation*}
    [\mathcal{A}]_{ij} =
    \begin{cases}
       a_{ij}, & \text{if } e_{i,j} \in \mathcal{E}, \\
       0, & \text{otherwise}.
    \end{cases}
\end{equation*}

The graph Laplacian is constructed as $L = D - \mathcal{A}$, where $\delta_i = \sum_{j=1}^N a_{ij}$ are the node degrees and $D = \mathrm{diag}(\delta)$ is the degree matrix. The Laplacian encodes important structural properties of the graph~\cite{yang2010decentralized}. In particular, $L \mathbf{1} = \mathbf{0}$, so the smallest eigenvalue of $L$ is always $0$. The second smallest eigenvalue is known as the \emph{algebraic connectivity} (or Fiedler value). It satisfies $\lambda_2 > 0$ if and only if the graph $\mathcal{G}$ is connected. Intuitively, $\lambda_2$ provides a quantitative measure of the robustness of connectivity: larger values correspond to better-connected networks, while $\lambda_2 = 0$ indicates that the network is disconnected.

In practice, it is often beneficial to use a smooth, differentiable weighting function which decreases as the inter-robot distance increases. Following~\cite{gasparri2017bounded}, we use
\begin{equation*}
    a_{ij} = 
    \begin{cases}
        e^{\tfrac{(R^2 - d_{i, j}^2)^2}{\varsigma}} - 1, & \text{if } d_{i, j} \leq R, \\
        0, & \text{otherwise},
    \end{cases}
\end{equation*}
where $\varsigma > 0$ is a tuning parameter to set the edge weight $a_{i,j} \leq 1$. We can also obtain
\begin{equation}
\label{eq:prel-grad-lambda}
    \nabla_{\mathbf{p}_i} \lambda_2(\mathbf{p}) \;=\; 
    \sum_{j}
    \nabla_{\mathbf{p}_i} a_{ij} \, (q_i - q_j)^2,
\end{equation}
where the summation is over all robots $j \neq i$, and $\mathbf{q} = [q_1,\ldots,q_N]^T$ is the eigenvector associated with $\lambda_2$. The gradient $\nabla_{\mathbf{p}_i} a_{ij}$ is given by
\begin{equation*}
    \nabla_{\mathbf{p}_i} a_{ij} = -\frac{2a_{ij}}{\varsigma} \, (R^2 - d_{i,j}^2)(\mathbf{p}_i - \mathbf{p}_j).
\end{equation*}

\subsection{B\'ezier Curve}
B\'ezier curves provide a convenient parametrization for smooth trajectory generation and have been widely adopted in motion planning. A B\'ezier curve of degree $n$ and duration $\tau$ is defined by a set of $n+1$ control points $\boldsymbol{\mathcal{P}}^{(m)} = \{\mathcal{P}_0^{(m)}, \dots, \mathcal{P}_n^{(m)}\} $ and can be expressed as
\begin{equation*}
    \mathcal{B}^{(m)}(t) = \sum_{j=0}^{n} b_{j, n}(t) \mathcal{P}_j^{(m)}, \quad t \in [0,\tau_m],
\end{equation*}
where $\tau_m > 0$ is the curve duration and $b_{j, n}(\cdot)$ are the Bernstein basis polynomials
\begin{equation*}
    b_{j, n}(t) = \binom{n}{j} \left(\frac{t}{\tau_m}\right)^j \left(1-\frac{t}{\tau_m}\right)^{n-j}.
\end{equation*}
B\'ezier curves are smooth by construction, and it is easy to evaluate their derivatives (which are also themselves B\'ezier curves). The resulting curve $\mathcal{B}(t)$ lies entirely within the convex hull of its control points. These properties make B\'ezier curves particularly well suited for trajectory optimization.

To generate long-horizon trajectories, we can concatenate multiple B\'ezier curves to form a piecewise spline. With $M$ segments, the full spline, indexed by $m \in \{0, 1, \dots, M-1\}$, can be defined by the collection of control points
\begin{equation*}
    \boldsymbol{\mathcal{P}} = \{ \boldsymbol{\mathcal{P}}^{(0)}, \dots, \boldsymbol{\mathcal{P}}^{(M-1)} \}, \quad 
    \boldsymbol{\mathcal{P}}^{(m)} = \{
\mathcal{P}^{(m)}_0, \dots,  \mathcal{P}^{(m)}_n   \}.
\end{equation*}
Smoothness across segments is enforced by continuity constraints on shared control points, i.e., $\mathcal{P}_n^{(m)} = \mathcal{P}_0^{(m+1)}$, and matching derivatives up to continuity order $C$.
In our framework, the optimization variables can be reduced to the set of control points $\boldsymbol{\mathcal{P}}$ of the piecewise spline.

\section{PROBLEM STATEMENT}
We consider a homogeneous team of $N$ robots navigating in $2$D space. The robot state is written as $\mathbf{x}_i = [\,\mathbf{p}_i^\top, \mathbf{v}_i^\top\,]^\top \in \mathbb{R}^4$, where $\mathbf{p}_i, \mathbf{v}_i \in \mathbb{R}^2$ denote position and velocity. We use the following double-integrator model:
\begin{equation}
\dot{\mathbf{x}} = A \mathbf{x} + B \mathbf{u},
\label{eq:prel-double-integrator}
\end{equation}
where $A = [\mathbf{0}, \mathbf{I}; \mathbf{0}, \mathbf{0}] \in \mathbb{R}^{4 \times 4}$ and $B = [\mathbf{0}; \mathbf{I}] \in \mathbb{R}^{4 \times 2}$. Here, $\mathbf{0} \in \mathbb{R}^{2 \times 2}$ and $\mathbf{I} \in \mathbb{R}^{2 \times 2}$ are the zero and identity matrices, respectively. The control input $\mathbf{u}_i \in \mathbb{R}^2$ is the acceleration. The velocity and acceleration are bounded by $\mathbf{v}_{\min}, \mathbf{v}_{\max} \in \mathbb{R}^2$ and $\mathbf{a}_{\min}, \mathbf{a}_{\max} \in \mathbb{R}^2$, respectively.
We denote the neighbors of robot $i$ by $\mathcal{N}_i = \{\, j \in \{1,2,\dots,N\} \mid j \neq i \,\}$. The state of all robots is $\mathbf{\chi} = [\,\mathbf{x}_1^\top, \dots , \mathbf{x}_N^\top\,]^\top \in \mathbb{R} ^{4N}$. We use $\mathbf{\xi} = [\,\mathbf{p}_1^\top, \dots, \mathbf{p}_N^\top\,]^\top$ to represent all robot positions.

The multi-robot system is represented by a time-varying graph $\mathcal{G}(\mathbf{\chi})$, where an edge is defined between two robots if their distance is within $R$.
The connectivity is preserved by maintaining the graph algebraic connectivity $\lambda_2(\mathbf{\chi}) > 0$. 

We assume that each robot has access to the global state and solves the motion-planning optimization problem locally. Our objective is to generate trajectories and control inputs concurrently for each robot, represented by a B\'ezier curve that:
\begin{enumerate}
    \item reaches goals without colliding with robots and obstacles,
    \item respects the initial state, smoothness requirements, and the dynamics model,
    \item maintains connectivity and recovers it when initially or temporarily lost. %
\end{enumerate}

\section{MPC--CLF--CBF Framework}
We propose an optimization-based trajectory generation algorithm, constrained by HOCLFs and HOCBFs, that promotes connectivity maintenance and recovery. We formulate connectivity preservation and collision avoidance as HOCBF constraints, and use HOCLF constraints to promote neighbor proximity; these recovery terms are emphasized when connectivity is lost.
Our approach optimizes a B\'ezier curve that generates a trajectory and the corresponding control inputs concurrently over a finite horizon. %

We first describe the continuous-time HOCLF/HOCBF constraints in Sec.~\ref{sec:method-constraints}, then present the Quadratic Programming (QP) formulation, gate slack penalties, and an SQP-based solution in Sec.~\ref{sec:qp-assembly}.

\subsection{Continuous-time Constraints}
\label{sec:method-constraints}
\subsubsection{Connectivity via HOCBF}
\label{sec:method-conn-hocbf}

To formulate connectivity maintenance, we define the following CBF~\cite{capelli2021decentralized}:
\begin{equation}
\label{eq:conn-hocbf}
    h^{\mathrm{conn}}(\mathbf{\chi}) = \lambda_2(\mathbf{\xi}) - \epsilon,
\end{equation}
where $\epsilon > 0$ is a minimum connectivity threshold. 
The relative degree of $h^{\mathrm{conn}}$ is two with respect to the model in~(\ref{eq:prel-double-integrator}). 
For robot $i$, using the expression for $\nabla_{\mathbf{p}_i} \lambda_2(\mathbf{p})$ in~\eqref{eq:prel-grad-lambda}, we can write its HOCBF inequality in affine form:
\begin{equation*}
    \nabla_{\mathbf{p}_i} \lambda_2(\mathbf{\xi})^\top \mathbf{u}_i + b_i^{\mathrm{conn}}(\mathbf{\chi}) \geq 0,
\end{equation*}
where $b_i^{\mathrm{conn}}(\mathbf{\chi}) = L_f^2 h^{\mathrm{conn}}(\mathbf{\chi}) + \alpha_2(\dot h^{\mathrm{conn}}(\mathbf{\chi}) + \alpha_1(h^{\mathrm{conn}}(\mathbf{\chi})))$ collects the terms independent of $\mathbf{u}_i$. We adopt linear class-$\mathcal{K}$ functions $\alpha_1(s) = k_1s$ and $\alpha_2(s) = k_2s$.

\begin{remark}
We assume that the Laplacian matrix has a simple Fiedler eigenvalue and $\nabla_\mathbf{\xi} \lambda_2(\mathbf{\xi})\neq0$. Small perturbations from noise or robot motion generally break perfectly symmetric configurations; under this assumption, the HOCBF constraint remains well-defined and enforceable.
\end{remark}

\subsubsection{Connectivity recovery via HOCLF}
As discussed in Section~\ref{sec:prel-conn}, $\lambda_2$ remains $0$ as long as connectivity is lost. The CBF controller alone cannot restore fleet connectivity once it is lost. We propose the following control Lyapunov function for robot pairs:
\begin{equation*}
V_{i,j}(\mathbf{x}) \;=\; w_{ij}\, \varphi\big(\|\mathbf{p}_i-\mathbf{p}_j\| - R\big),
\end{equation*}
where $\mathbf{p}_i$ and $\mathbf{p}_j$ denote the positions of the current robot and its neighbor $j\in \mathcal{N}_i$, respectively, and $w_{ij} = \rho^{\,d_{ij}/R} > 0$ is a distance-decayed weight with decay rate $\rho \in (0, 1)$, assigning a higher penalty to closer neighbors.
We define the one-sided penalty function as $\varphi(z)=\left(\max\left(0,\,z\right)\right)^{2}$, so the recovery penalty is active only when $\|\mathbf{p}_i-\mathbf{p}_j\|>R$.
The relative degree of $V_{i,j}(\mathbf{x})$ is two with respect to the model in~(\ref{eq:prel-double-integrator}). Expanding~\eqref{eq:prel-hoclf}, the HOCLF condition can be written as follows:
\begin{equation*}
    L_f^2 V_{i,j}(\mathbf{x}) + L_g L_f V_{i,j}(\mathbf{x})\, \mathbf{u} + \alpha_2\!\big(\dot V_{i,j}(\mathbf{x}) + \alpha_1(V_{i,j}(\mathbf{x}))\big) \leq 0,
\end{equation*}
or in affine form:
\begin{equation*}
    L_g L_f V_{i,j}(\mathbf{x})\,\mathbf{u} + b^{\mathrm{clf}}_{i,j}(\mathbf{x}) \leq 0.
\end{equation*}
The HOCLF drives pairwise robot distances within the connectivity radius $R$, while the HOCBF condition promotes $\lambda_2 \ge \epsilon$.

\subsubsection{Collision avoidance}
We use the CBF from~\cite{egerstedt2018robot}, which maintains a minimum safety distance for robot--robot and robot--obstacle pairs:
\begin{equation*}
h^{\mathrm{safe}}_{i,o}(\mathbf{x}) = \|\mathbf{p}_i-\mathbf{o}\|^2 - d_{\min}^2, 
\quad \forall\, i,\ \forall\, \mathbf{o}\in\mathcal{O}_i,
\end{equation*}
where $\mathcal{O}_i=\{\mathbf{p}_j:\, j \in \mathcal{N}_i\}\cup\{\mathbf{o}_\ell\}$ collects the centers of neighboring robots and obstacles. Here, $\mathbf{o}_\ell \in \mathbb{R}^2$ is the center of each obstacle, and $d_{\min}$ is the minimum safety distance. The relative degree of $h^{\mathrm{safe}}_{i,o}(\mathbf{x})$ is two with respect to the model in~(\ref{eq:prel-double-integrator}). As in Sec.~\ref{sec:method-conn-hocbf}, this formulation yields linear constraints on $\mathbf{u}$.

\subsection{Solving Optimization}
\label{sec:qp-assembly}
\subsubsection{QP formulation}
We formulate trajectory generation as a QP subject to the aforementioned connectivity-preserving and collision-avoidance constraints, minimizing goal deviation and control effort over a planning horizon.
Note that both the HOCLF and HOCBF constraints are defined in continuous time over the planning horizon, which introduces an infinite number of constraints and makes the problem intractable. Inspired by~\cite{11235958}, we sample the HOCLF and HOCBF constraints at discrete time steps over the horizon to obtain a computationally tractable approximation, although formal continuous-time guarantees do not directly hold.
At a replanning time $t_0$, the system predicts a continuous-time B\'ezier trajectory over a horizon $\tau = K\sigma$, corresponding to $K$ steps of duration $\sigma$.
Additionally, since the output trajectories are parameterized as piecewise B\'ezier curves, the robots' velocities and accelerations can be directly obtained by evaluating their derivatives, and the double-integrator dynamics in \eqref{eq:prel-double-integrator} are respected.
At each time step, we solve the following QP for each robot $i$, where $j \in \mathcal{N}_i$:
\begin{subequations}
\begin{IEEEeqnarray}{lCl}
\IEEEeqnarraymulticol{3}{l}{\argmin_{\boldsymbol{\mathcal{P}}} \quad \mathcal{J}_{\mathrm{goal}} + \mathcal{J}_{\mathrm{effort}} 
+ \mathcal{J}_{\mathrm{slack}}} \label{QPcost}\\
\text{s.t.} 
&~& \frac{d^c \mathcal{B}^{(0)}(0)}{dt^c} = \left.\frac{d^c \mathbf{p}_i(t)}{dt^c}\right|_{t=t_0}, \forall c \in \{0,\dots,C\}
 \label{QPconst:init_state}\\
&~& \begin{aligned} \frac{d^c \mathcal{B}^{(m)}(\tau_m)}{d t^c}=\frac{d^c \mathcal{B}^{(m + 1)}(0)}{d t^c}, ~&\forall c \in \{0,\dots,C\} \\ &\forall m \in\{0, \ldots, M\!-\!2\} \end{aligned} \label{QPconst:continuity}\\
&~& A^{\mathrm{safe}}_{i, o}\,\hat{\mathbf u}_k + b^{\mathrm{safe}}_{i, o} \geq 0, ~\forall \mathbf{o} \in \mathcal{O}_{i} \label{QPconst:safety}\\
&~& A^{\mathrm{conn}}_{i}\,\hat{\mathbf u}_k + b^{\mathrm{conn}}_{i} \geq -\,\varepsilon^{\mathrm{conn}}_{i} \label{QPconst:conn}\\
&~& A^{\mathrm{clf}}_{i, j}\,\hat{\mathbf u}_k + b^{\mathrm{clf}}_{i, j} \leq \varepsilon^{\mathrm{clf}}_{i, j}, ~\forall j \in \mathcal{N}_{i} \label{QPconst:rec}\\
&~& \mathbf{v}_{\min} \preceq \hat{\mathbf v}_k \preceq \mathbf{v}_{\max}, ~\forall k \in \{0,\dots,K-1\} \label{QPconst:vel_limits}\\
&~& \mathbf{a}_{\min} \preceq \hat{\mathbf u}_k \preceq \mathbf{a}_{\max}, ~\forall k \in \{0,\dots,K-1\} \label{QPconst:acc_limits}\\
&~& \varepsilon^{\mathrm{conn}}_{i} \geq 0,\ \varepsilon^{\mathrm{clf}}_{i,j} \geq 0,\ \forall j \in \mathcal{N}_{i} \label{QPconst:slacks}
\end{IEEEeqnarray}
\label{QPconst}
\end{subequations}%
\noindent where $\preceq$ denotes element-wise inequality, and $C$ is the highest derivative order required for continuity.
Specifically, \eqref{QPconst:init_state} ensures the initial condition, \eqref{QPconst:continuity} ensures continuity between two piecewise B\'ezier curves, and \eqref{QPconst:vel_limits} and \eqref{QPconst:acc_limits} are the physical limits.

We consider two objectives in the optimization. The goal-reaching cost penalizes the deviation of the predicted output from the desired goal,
$\mathcal{J}_{\mathrm{goal}} = \sum_{k = 0}^{K-1} \omega_{k}\left\Vert \hat{\mathbf{y}}_{k} - \mathbf{y}_{\mathrm{desired}}\right\Vert_{2}^{2},$
while the control effort cost penalizes derivatives of the trajectory,
$\mathcal{J}_{\mathrm{effort}} = \sum_{c=1}^{C} \theta_{c} \int_{t_{0}}^{t_{0}+\tau} \left\Vert\frac{d^{c}}{dt^{c}} \mathcal{B}(t;\boldsymbol{\mathcal{P}})\right\Vert_{2}^{2} \, dt$.
Both costs are quadratic in the decision variable $\boldsymbol{\mathcal{P}}$.

\subsubsection{Smooth weighting via gate slack penalties}
To retain feasibility under competing objectives, we introduce nonnegative slack variables $\varepsilon^{\mathrm{conn}}_{i}$ and $\varepsilon^{\mathrm{clf}}_{i,j}$, which relax the connectivity-maintenance HOCBF constraint~\eqref{QPconst:conn} and the recovery HOCLF constraint~\eqref{QPconst:rec}, respectively.

We smoothly reweight the slack penalties according to the current algebraic connectivity. We define a gate function
\begin{equation}
\eta(\lambda_2)=\eta_{\min} + \tfrac{1}{2}(1-2\eta_{\min}) \left(1+\tanh\left(\frac{\lambda_2-\epsilon}{w_{\eta}}\right)\right).
\label{eq:gate_function}
\end{equation}
Here, $w_{\eta}>0$ is the gate steepness parameter that controls the transition near the connectivity threshold $\epsilon$.
By construction, $\eta(\lambda_2)\in[\eta_{\min},\,1-\eta_{\min}]$ for a small $\eta_{\min}>0$.

We then define a slack cost,
\begin{multline*}
\mathcal{J}_{\mathrm{slack}} 
= \eta(\lambda_2)\sum_{i} \gamma_i\big(\varepsilon^{\mathrm{conn}}_{i}\big)^2 \\
+ (1-\eta(\lambda_2))\sum_{i}\sum_{j\in\mathcal{N}_i} \mu_{ij}\big(\varepsilon^{\mathrm{clf}}_{i,j}\big)^2,
\end{multline*}
where $\gamma_i,\mu_{ij}>0$ are slack weights. Both constraints remain present in the QP; the gate only reweights the quadratic slack penalties: below the connectivity threshold $\epsilon$, the gate emphasizes recovery by assigning a higher penalty to $\varepsilon^{\mathrm{clf}}_{i,j}$, while above the threshold, the gate emphasizes connectivity maintenance by assigning a higher penalty to $\varepsilon^{\mathrm{conn}}_{i}$.

\subsubsection{Solution via SQP}
\label{sec:method-sqp}
Since the connectivity constraints~\eqref{eq:conn-hocbf} are nonlinear with respect to $\chi$, we linearize them around the most recent predicted trajectory to obtain affine approximations.
We solve~\eqref{QPconst} with SQP under a fixed number of iterations $L$. The predicted position, velocity, and control at step $k \in \{0,\dots,K-1\}$ are denoted $\hat{\mathbf{y}}_{k}$, $\hat{\mathbf{v}}_{k}$, and $\hat{\mathbf{u}}_{k}$, and are evaluated from the optimized trajectory. Initially ($l=0$), the QP is solved using only the current state, yielding $\prescript{0}{}{\hat{\mathbf{y}}_k}$, $\prescript{0}{}{\hat{\mathbf{v}}_k}$, and $\prescript{0}{}{\hat{\mathbf{u}}_k}$. For subsequent iterations $l=1, \dots, L-1$, the predicted states from the previous iteration are independent of decision variables and treated as constants, i.e., 
\begin{equation*}
    A (\prescript{l-1}{}{\hat{\mathbf{y}}_k},\prescript{l-1}{}{\hat{\mathbf{v}}_k})  \prescript{l}{}{\hat{\mathbf{u}}_k} + b(\prescript{l-1}{}{\hat{\mathbf{y}}_k},\prescript{l-1}{}{\hat{\mathbf{v}}_k}) 
\end{equation*}
as the LHS of the constraints in~\eqref{QPconst}.
This iterative procedure ensures that, despite the nonlinearity of the CLF--CBF constraints, each iteration is a convex QP that can be solved efficiently using off-the-shelf QP solvers.

\begin{figure}[!hbtp]
    \centering
    \begin{subfigure}[b]{\linewidth}
        \centering
        \hspace{0.06\linewidth}
        \includegraphics[height=0.34\linewidth,keepaspectratio]{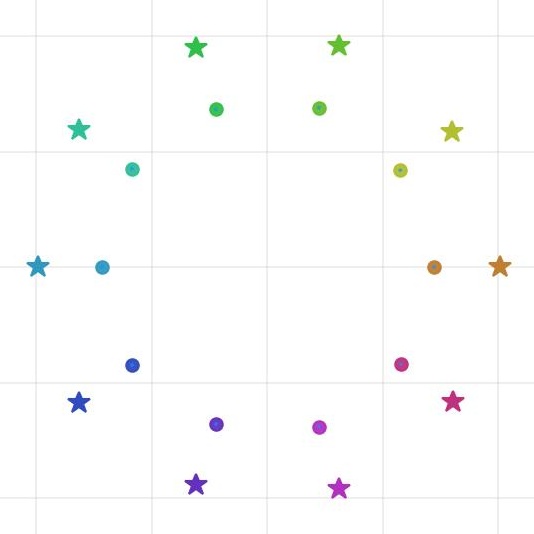}
        \hspace{0.04\linewidth}
    \includegraphics[height=0.34\linewidth,keepaspectratio]{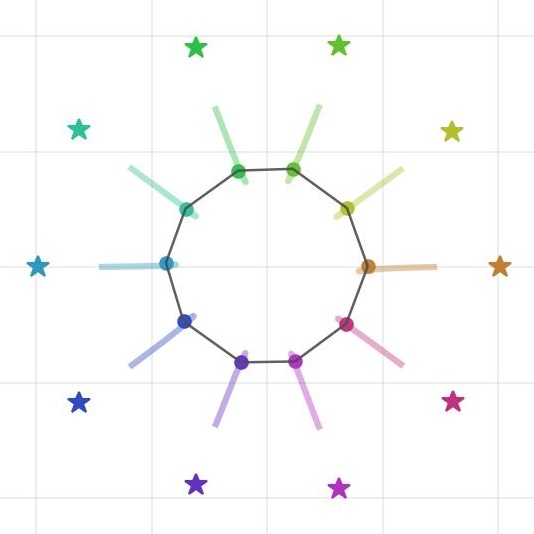}
        \hspace{0.22\linewidth}
        \\[3pt]
        \begin{subfigure}[b]{\linewidth}
            \centering
            \includegraphics[width=0.9\linewidth]{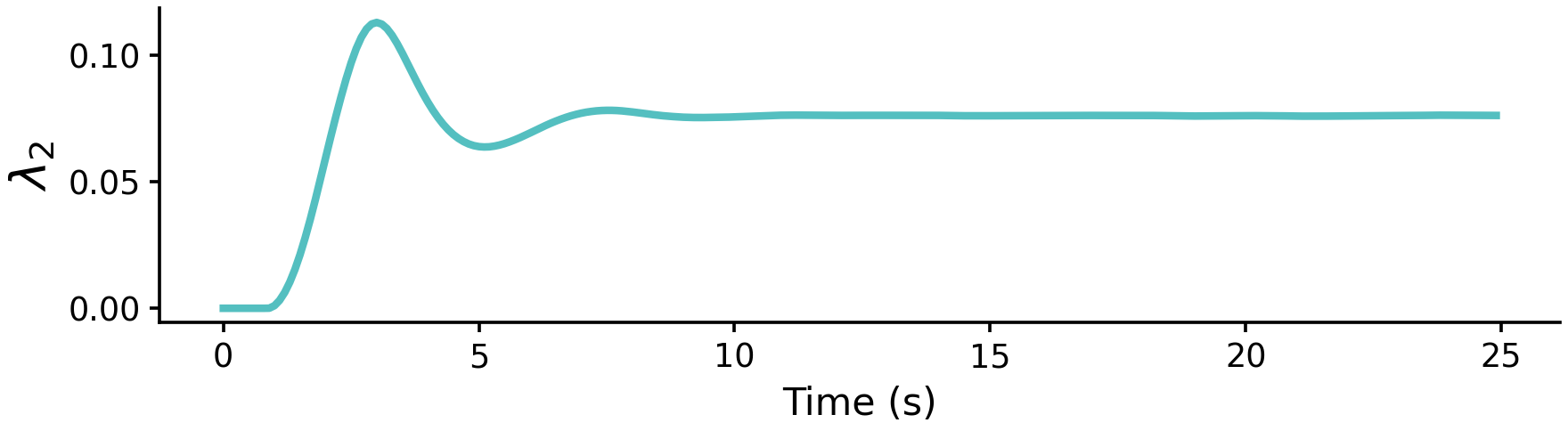}
            \label{fig:recovery-lambda2}
        \end{subfigure}
        \caption{Case 1: snapshots at $t=\SI{0}{s}$ and $t=\SI{5}{s}$, and $\lambda_2$ over time.}
        \label{fig:recovery-case1}
    \end{subfigure}

    \vspace{0.5em}

    \begin{subfigure}[b]{\linewidth}
        \centering
        \hfill
        \begin{subfigure}[b]{0.48\linewidth}
            \centering
            \includegraphics[width=\linewidth]{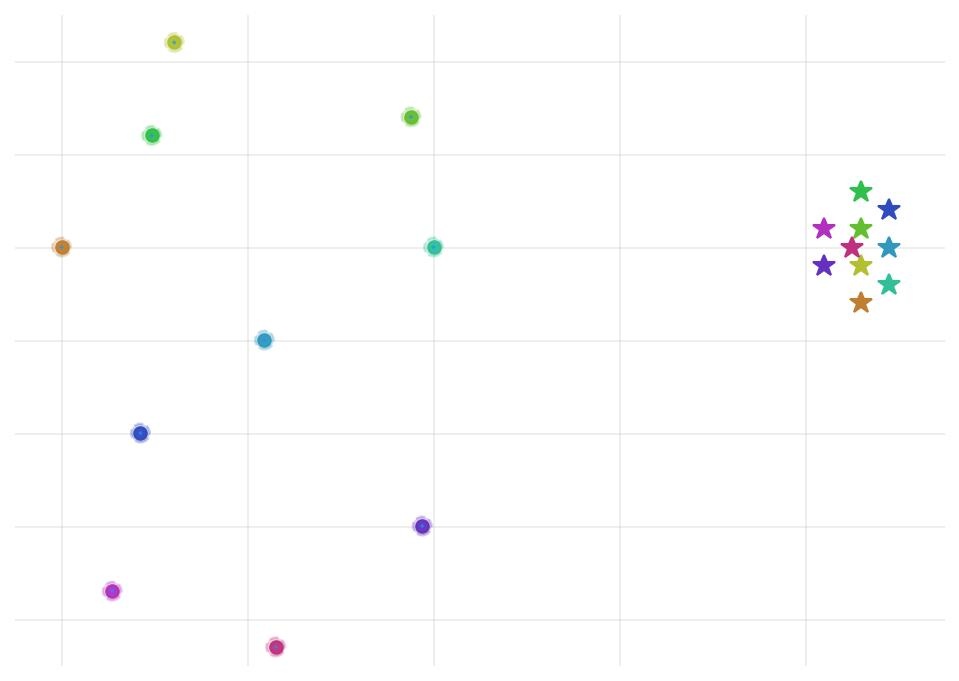}
        \end{subfigure}
        \hfill
        \begin{subfigure}[b]{0.48\linewidth}
            \centering
            \includegraphics[width=\linewidth]{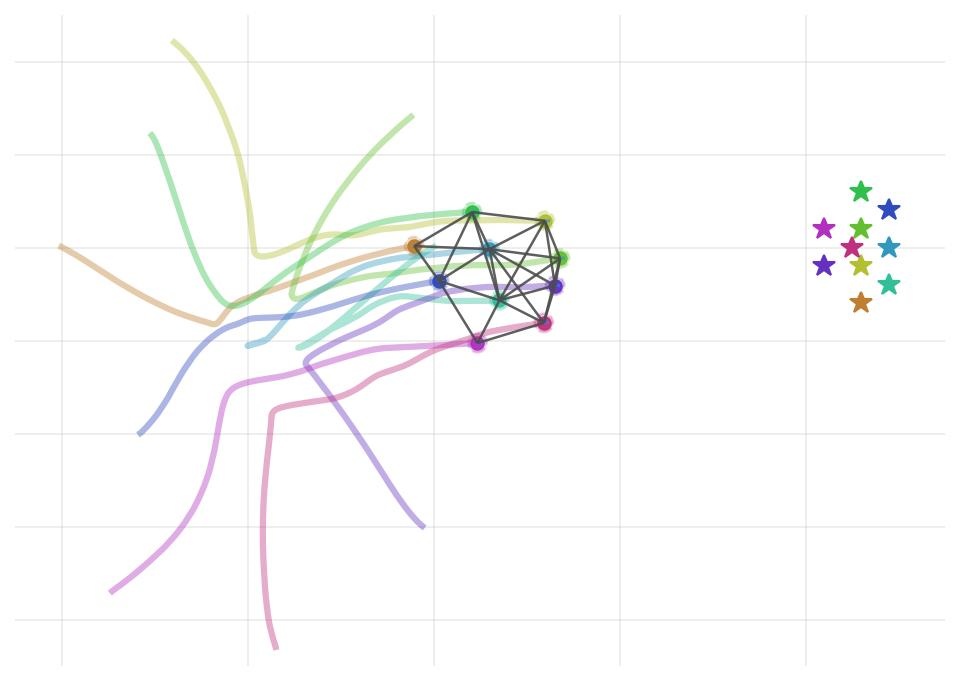}
        \end{subfigure}
        \hfill
        \\[3pt]
        \begin{subfigure}[b]{\linewidth}
            \centering
            \includegraphics[width=0.9\linewidth]{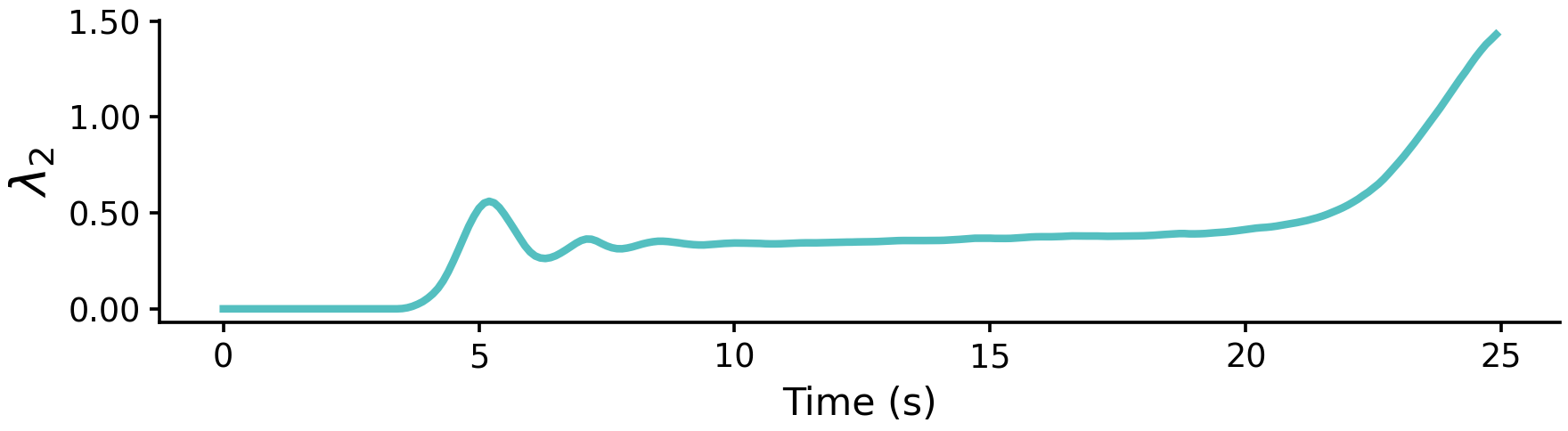}
            \label{fig:recovery2-lambda2}
        \end{subfigure}
        \caption{Case 2: snapshots at $t=\SI{0}{s}$ and $t=\SI{12}{s}$, and $\lambda_2$ over time.}
        \label{fig:recovery-case2}
    \end{subfigure}

    \caption{Snapshots of robot trajectories during disconnection and subsequent connectivity recovery. The stars indicate robot goal positions, and the black lines indicate communication links between neighboring robots.}
    \label{fig:recovery-sequence}
\end{figure}

\section{SIMULATION EXPERIMENTS}
We implement the algorithm in C++ with the CPLEX QP solver. For all instances, we use piecewise splines with $M = 4$ segments, where each B\'ezier curve has degree $3$ and duration $\tau = \SI{1}{s}$. We set $\epsilon = 0.1$ as the algebraic connectivity threshold.
As mentioned in Section~\ref{sec:qp-assembly}, we sample the HOCBF and HOCLF constraints at discrete time steps to approximate the continuous-time conditions.
We conducted a parameter search and found that $\sigma=\SI{0.1}{s}$ and a control frequency of $\SI{100}{Hz}$ achieve a good balance between constraint satisfaction and real-time performance. Additionally, we set the gate steepness parameter $w_{\eta} = 0.01$, except in Section~\ref{sec:gate_steepness}, and the minimum gate value $\eta_{\min} = 0.001$ in all experiments.
For experiments that involve obstacles, we model circular obstacles with varying radii. The obstacle density is defined as the fraction of the obstacle region's area that is occupied by obstacles.

We evaluate the efficacy of our MPC--CLF--CBF framework by benchmarking against two baseline frameworks: 1) \emph{CLF--CBF} based on~\cite{capelli2020connectivity}, augmented with the same CLF recovery constraints used in our method; 2) \emph{MPC--CBF}~\cite{11235958}, which uses the same trajectory generation mechanism and connectivity HOCBF constraints but omits HOCLF recovery terms.

We use the following metrics in the quantitative experiments:
\begin{itemize}
    \item \textbf{Success Rate}: the percentage of robots that reach their goals without collisions.
    \item \textbf{Makespan}: the total time required for all robots to complete the task.
    \item \textbf{Percentage Connected}: the fraction of time for which the communication graph is topologically connected, i.e., $\lambda_2 > 0$. This differs from satisfying the prescribed connectivity margin $\lambda_2 \geq \epsilon$.
\end{itemize}

\subsection{Connectivity Recovery}
We demonstrate our motion planner's connectivity-recovery capability through qualitative experiments. We consider two navigation tasks with $10$ initially disconnected robots, i.e., $\lambda_2=0$ at $t=0$, as shown in Fig.~\ref{fig:recovery-sequence}.
The CLF constraints reduce the pairwise distances between robots until the robots become connected.

In the first example, the desired goal configuration, which is also shown as the initial configuration in Fig.~\ref{fig:recovery-case1}, is disconnected, but our motion planner yields a reachable configuration that satisfies the connectivity constraints.
The robots reduce their pairwise distances to establish connectivity, even though that requires deviating from their goals.
In the second example, the goal configuration is connected. As shown in Fig.~\ref{fig:recovery-case2}, the robots recover and maintain connectivity while proceeding toward their goals, leading to a sustained increase in $\lambda_2$.

As connectivity improves, the gate $\eta(\lambda_2)$ increases smoothly, leading to a smooth tightening of the connectivity HOCBF constraints and relaxation of HOCLF constraints. 

Therefore, our motion planner demonstrates the ability to recover connectivity during navigation and before the robots reach their goals in both connected and disconnected goal configurations.

\subsection{Gate Steepness Parameter Ablation}
\label{sec:gate_steepness}
We quantitatively evaluate the influence of the gate steepness parameter $w_{\eta}$ in~\eqref{eq:gate_function} on our planner. The gate steepness parameter $w_{\eta}$ controls the sensitivity of our planner's trade-off between the connectivity maintenance HOCBFs and the recovery HOCLFs. %
In particular, a smaller $w_\eta$ results in a more responsive switch, i.e., a rapid change in $\eta({\lambda_2)}$, when $\lambda_{2}$ is near the threshold $\epsilon$, while a larger value leads to a slower switch. %

\begin{table}[!h]
\centering
\footnotesize
\setlength{\tabcolsep}{0.5pt}
\renewcommand{\arraystretch}{0.5}

\begin{tabular}{lccc}
\toprule
Method & \shortstack{Worst Conn.\\(\%) $\uparrow$} & \shortstack{Mean Conn.\\(\%) $\uparrow$} & \shortstack{Runtime\\(ms) $\downarrow$} \\
\midrule
No-gating              & 33.00 & 81.27 & 74.38 \\
\midrule
$w_{\eta}=0.001$  & 62.95 & 95.57 & 62.66 \\
\midrule
$w_{\eta}=0.01$   & \textbf{70.77} & \textbf{96.71} & \textbf{62.47} \\
\midrule
$w_{\eta}=0.1$    & 51.27 & 93.53 & 72.24 \\
\bottomrule
\end{tabular}
\caption{Gating ablation for $10$ robots at $20\%$ obstacle density over $10$ different robot and obstacle configurations. We report the worst-case percentage connected across configurations, the mean percentage connected across configurations, and the mean optimization solver runtime.}
\label{tab:gating_width}
\end{table}

As shown in Table~\ref{tab:gating_width}, introducing the
gate function improves both worst-case and overall connectivity without increasing the optimization solver runtime.
Without the gate function, fixed coefficients are used for HOCBF and HOCLF constraints, which impose competing constraints and reduce connectivity performance. 
With the gate function, a large $w_\eta$ produces more gradual reweighting as connectivity decreases, so the recovery penalty becomes dominant more slowly. Consequently, we observe robots remaining separated longer before recovery.
Since different $w_\eta$ values yield similar runtime performance, we adopt the intermediate $w_{\eta} = 0.01$ for all experiments.

\subsection{Benchmarking Quantitative Results} 

\begin{table*}[t]
\centering
\setlength{\tabcolsep}{3.5pt}
\resizebox{\textwidth}{!}{%
\renewcommand{\arraystretch}{1.0}%
\begin{tabular}{cccccccccc}
\toprule
\multirow{2}{*}{\shortstack{Number of\\robots}} & \multicolumn{3}{c}{Success Rate (\%) $\uparrow$} & \multicolumn{3}{c}{Percentage Connected (\%) $\uparrow$} & \multicolumn{3}{c}{Makespan (s) $\downarrow$} \\
\cmidrule(lr){2-4} \cmidrule(lr){5-7} \cmidrule(lr){8-10}
& MPC--CLF--CBF  & CLF--CBF & MPC--CBF & MPC--CLF--CBF & CLF--CBF & MPC--CBF & MPC--CLF--CBF & CLF--CBF & MPC--CBF \\
\midrule
\multicolumn{10}{c}{\textbf{Obstacle Density 0\%}} \\
\midrule
4 & $100.0 \pm 0.0$ & $100.0 \pm 0.0$ & $100.0 \pm 0.0$ & $100.0 \pm 0.0$ & $100.0 \pm 0.0$ & $100.0 \pm 0.0$ & $31.5 \pm 2.0$ & $33.4 \pm 1.9$ & $\mathbf{30.6 \pm 1.6}$ \\
6 & $96.7 \pm 7.0$ & $96.7 \pm 7.0$ & $96.7 \pm 7.0$ & $100.0 \pm 0.0$ & $100.0 \pm 0.0$ & $100.0 \pm 0.0$ & $\mathbf{39.3 \pm 11.1}$ & $42.0 \pm 9.7$ & $39.8 \pm 10.8$ \\
8 & $96.2 \pm 8.4$ & $96.2 \pm 8.4$ & $96.2 \pm 8.4$ & $100.0 \pm 0.0$ & $100.0 \pm 0.0$ & $100.0 \pm 0.0$ & $40.2 \pm 10.9$ & $43.7 \pm 8.8$ & $40.2 \pm 10.8$ \\
10 & $98.0 \pm 4.2$ & $98.0 \pm 4.2$ & $97.0 \pm 6.8$ & $100.0 \pm 0.0$ & $100.0 \pm 0.0$ & $100.0 \pm 0.0$ & $41.9 \pm 9.9$ & $41.9 \pm 6.8$ & $41.8 \pm 9.8$ \\
12 & $88.3 \pm 10.5$ & $84.2 \pm 14.9$ & $88.3 \pm 10.5$ & $100.0 \pm 0.0$ & $100.0 \pm 0.0$ & $96.5 \pm 3.2$ & $56.4 \pm 7.8$ & $58.4 \pm 5.2$ & $\mathbf{55.8 \pm 6.9}$ \\
\midrule
\multicolumn{10}{c}{\textbf{Obstacle Density 10\%}} \\
\midrule
4 & $100.0 \pm 0.0$ & $100.0 \pm 0.0$ & $100.0 \pm 0.0$ & $\mathbf{100.0 \pm 0.0}$ & $98.1 \pm 4.6$ & $75.5 \pm 19.7$ & $34.0 \pm 3.7$ & $37.2 \pm 4.7$ & $\mathbf{33.5 \pm 3.2}$ \\
6 & $96.7 \pm 7.0$ & $\mathbf{98.3 \pm 5.3}$ & $96.7 \pm 7.0$ & $\mathbf{100.0 \pm 0.0}$ & $98.3 \pm 4.2$ & $80.0 \pm 11.8$ & $41.1 \pm 10.2$ & $43.8 \pm 6.6$ & $\mathbf{40.6 \pm 10.6}$ \\
8 & $\mathbf{97.5 \pm 5.3}$ & $96.2 \pm 6.0$ & $95.0 \pm 10.5$ & $\mathbf{100.0 \pm 0.0}$ & $91.8 \pm 4.2$ & $76.4 \pm 19.9$ & $42.0 \pm 11.2$ & $49.0 \pm 7.9$ & $\mathbf{41.8 \pm 9.7}$ \\
10 & $93.2 \pm 11.7$ & $89.7 \pm 5.8$ & $\mathbf{96.0 \pm 5.2}$ & $\mathbf{95.5 \pm 6.2}$ & $92.1 \pm 8.9$ & $88.3 \pm 18.6$ & $52.8 \pm 6.0$ & $55.0 \pm 3.4$ & $\mathbf{49.1 \pm 9.8}$ \\
12 & $\mathbf{97.8 \pm 14.9}$ & $94.0 \pm 9.1$ & $94.2 \pm 6.9$ & $\mathbf{95.8 \pm 10.1}$ & $91.9 \pm 4.5$ & $88.0 \pm 13.9$ & $\mathbf{50.1 \pm 3.7}$ & $54.6 \pm 7.7$ & $52.4 \pm 8.9$ \\

\midrule
\multicolumn{10}{c}{\textbf{Obstacle Density 20\%}} \\
\midrule
4 & $100.0 \pm 0.0$ & $95.0 \pm 15.8$ & $100.0 \pm 0.0$ & $\mathbf{100.0 \pm 0.0}$ & $93.1 \pm 5.0$ & $48.9 \pm 27.1$ & $38.0 \pm 3.3$ & $42.2 \pm 7.2$ & $37.4 \pm 3.1$ \\
6 & $\mathbf{96.7 \pm 7.0}$ & $95.0 \pm 8.1$ & $95.0 \pm 8.1$ & $\mathbf{100.0 \pm 0.0}$ & $90.7 \pm 2.8$ & $55.6 \pm 25.0$ & $\mathbf{47.9 \pm 8.6}$ & $50.7 \pm 7.0$ & $49.4 \pm 9.1$ \\
8 & $\mathbf{97.5 \pm 5.3}$ & $95.0 \pm 8.7$ & $96.2 \pm 6.0$ & $\mathbf{95.8 \pm 9.4}$ & $93.4 \pm 2.3$ & $60.6 \pm 20.8$ & $50.5 \pm 7.2$ & $53.5 \pm 5.4$ & $50.0 \pm 7.5$ \\
10 & $86.0 \pm 23.2$ & $88.0 \pm 31.2$ & $\mathbf{95.0 \pm 7.1}$ & $\mathbf{96.7 \pm 7.0}$ & $93.7 \pm 3.1$ & $61.3 \pm 26.0$ & $55.0 \pm 5.8$ & $57.3 \pm 2.8$ & $\mathbf{52.4 \pm 7.2}$ \\
12 & $\mathbf{88.3 \pm 11.2}$ & $86.7 \pm 11.2$ & $85.8 \pm 11.8$ & $\mathbf{96.4 \pm 7.2}$ & $95.1 \pm 3.0$ & $50.0 \pm 19.4$ & $57.0 \pm 5.0$ & $59.2 \pm 2.5$ & $56.8 \pm 5.3$ \\
\bottomrule
\end{tabular}%
}
\caption{Scalability results across obstacle densities as the team size increases from $4$ to $12$ robots. Each statistic is computed over $10$ different robot and obstacle configurations.}
\label{tab:scalability_table}
\end{table*}

\begin{figure*}[t]
    \centering
    \begin{subfigure}[b]{0.32\linewidth}
        \centering
        \includegraphics[width=\linewidth]{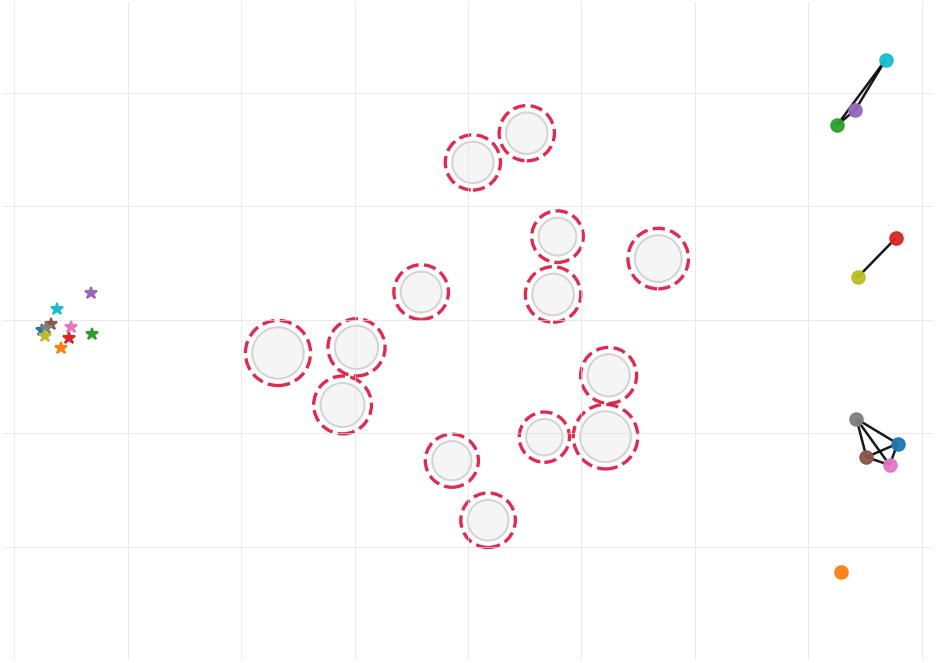}
        \caption{$t = \SI{0}{s}$}
        \label{fig:general-0}
    \end{subfigure}
    \hfill
    \begin{subfigure}[b]{0.32\linewidth}
        \centering
        \includegraphics[width=\linewidth]{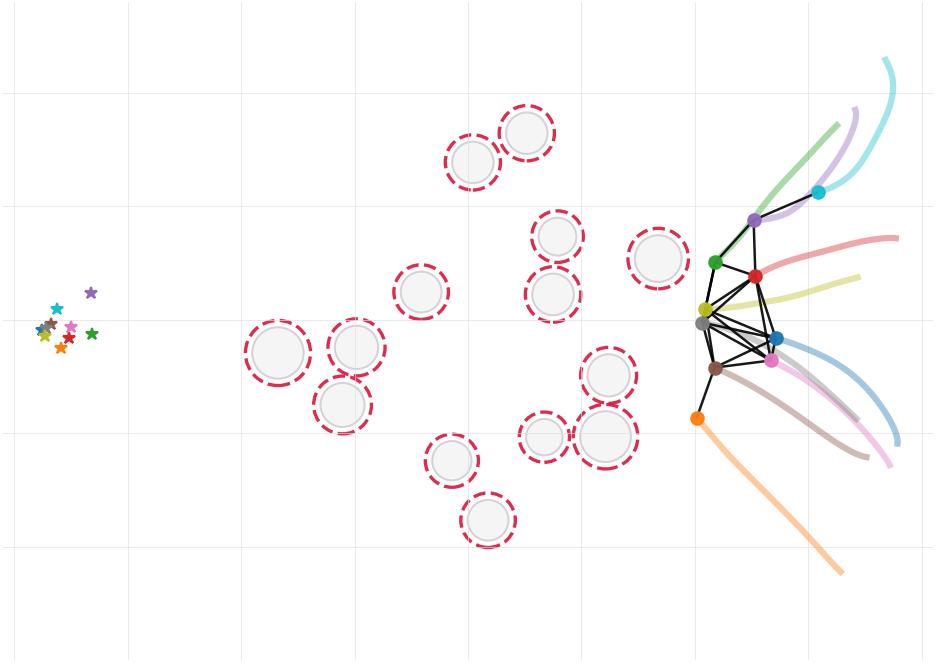}
        \caption{$t = \SI{5}{s}$}
        \label{fig:general-1}
    \end{subfigure}
    \hfill
    \begin{subfigure}[b]{0.32\linewidth}
        \centering
        \includegraphics[width=\linewidth]{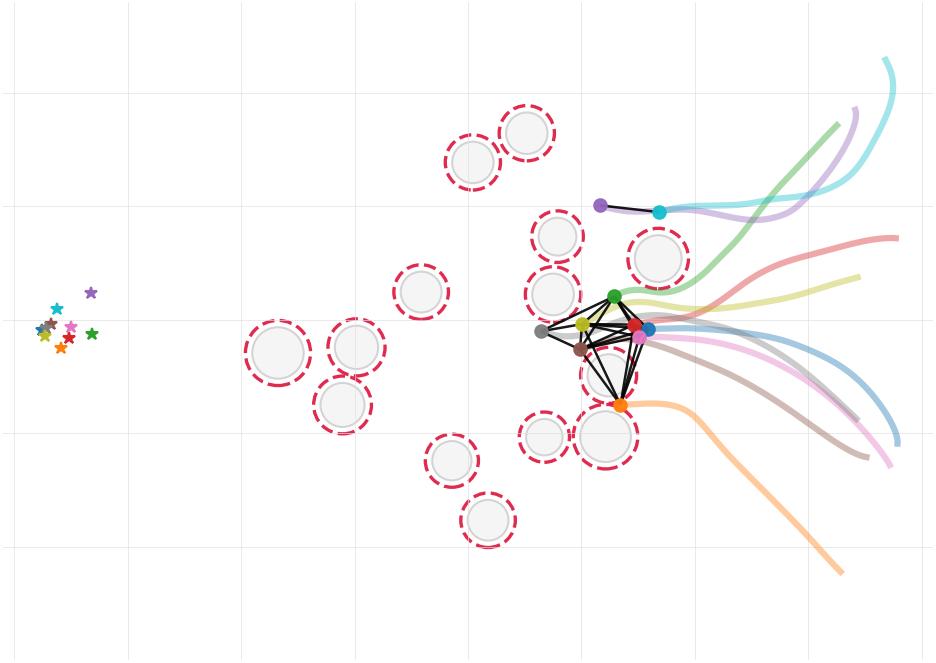}
        \caption{$t = \SI{10}{s}$}
        \label{fig:general-2}
    \end{subfigure}

    \vskip\baselineskip

    \begin{subfigure}[b]{0.32\linewidth}
        \centering
        \includegraphics[width=\linewidth]{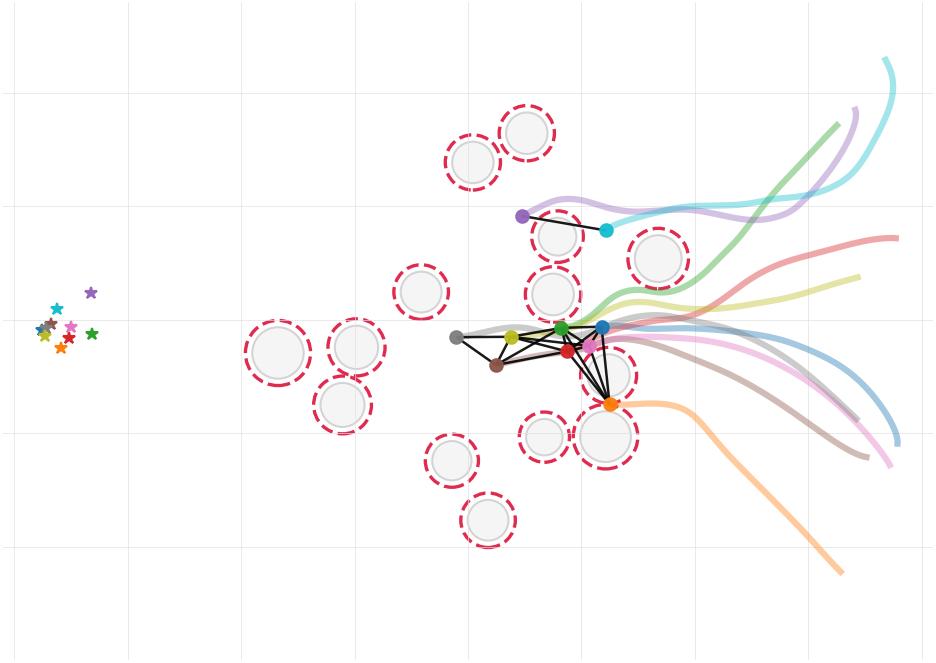}
        \caption{$t = \SI{12.5}{s}$}
        \label{fig:general-3}
    \end{subfigure}
    \hfill
        \begin{subfigure}[b]{0.32\linewidth}
        \centering
        \includegraphics[width=\linewidth]{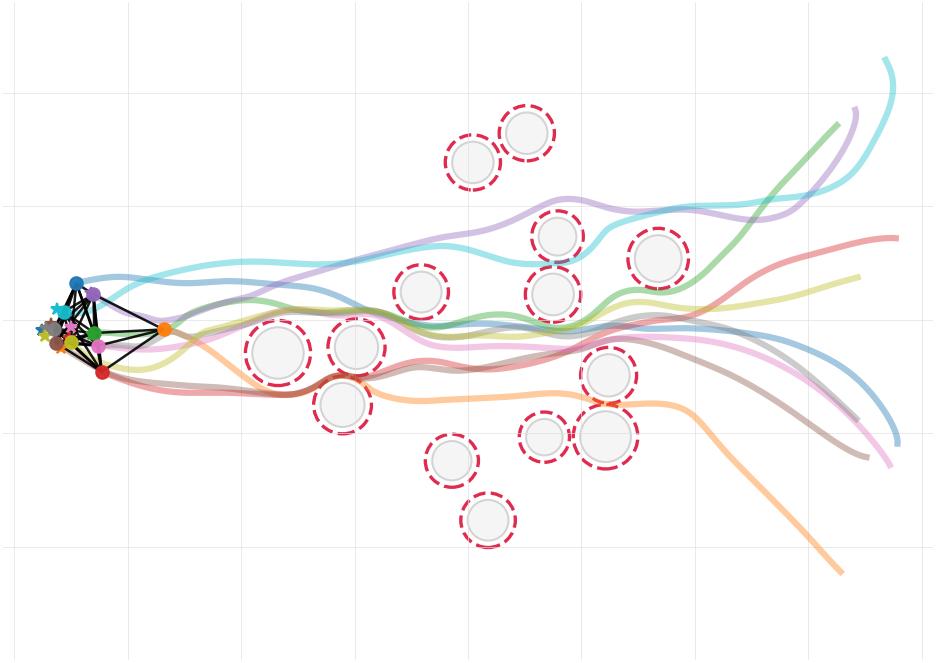}
        \caption{$t = \SI{35}{s}$}

        \label{fig:general-4}
    \end{subfigure}
    \hfill
    \begin{subfigure}[b]{0.32\linewidth}
        \centering
        \includegraphics[width=\linewidth]{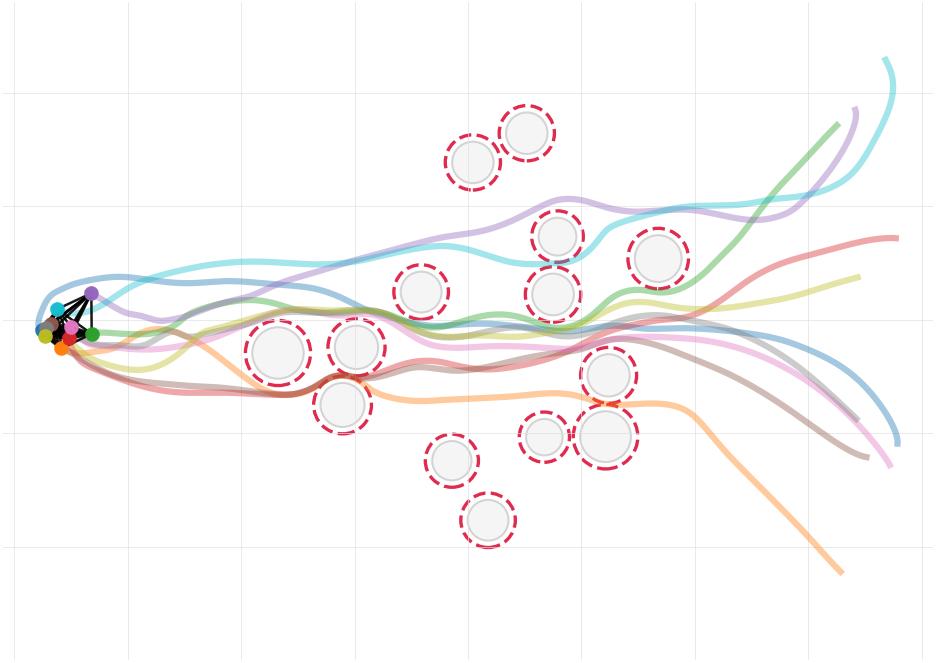}
        \caption{$t = \SI{54.2}{s}$}
        \label{fig:general-5}
    \end{subfigure}

    \caption{Illustrative trajectories of $10$ robots under the proposed \emph{MPC--CLF--CBF} controller in a $10\%$ obstacle-density instance. The team starts disconnected, reconnects, temporarily separates into two subgroups while navigating around obstacles, and recovers connectivity before the robots reach their goals. Obstacles and their safety margins are shown as gray disks and red dashed circles, respectively; robot goals are shown as stars, trajectories as colored lines, and active communication links as black lines.}
    \label{fig:obstacle_case}
\end{figure*}

We benchmark the performance of our \emph{MPC--CLF--CBF} against two baseline methods, namely, \emph{CLF--CBF} and \emph{MPC--CBF}, 
in a navigation task where robots start on one side of the workspace, cross the obstacle region, and reach their goals, as shown in Fig.~\ref{fig:obstacle_case}. We create instances with $4$ to $12$ robots and maps with different obstacle densities.
All experiments are conducted in a square workspace with position bounds $x, y \in \left[ \SI{-200}{m}, \SI{200}{m} \right]$. The velocity limits are set to $\left[ \SI{-15}{m/s}, \SI{15}{m/s} \right]$, and the acceleration limits are set to $\left[ \SI{-20}{m/s^2}, \SI{20}{m/s^2} \right]$. The minimum safety distance is set to $d_{\mathrm{min}} = \SI{2}{m}$ and the connectivity range is set to $R = \SI{40}{m}$. Each trial runs for $\SI{60}{s}$, and the task is considered complete when all robots come within $\SI{0.25}{m}$ of their goals. Obstacles are placed in the region $x, y \in \left[ \SI{-100}{m}, \SI{100}{m} \right]$.

Table~\ref{tab:scalability_table} benchmarks the performance of all three algorithms across different obstacle densities. The safety CBF constraint~\eqref{QPconst:safety} is a hard constraint, i.e., it has no slack variable. No robot--robot or robot--obstacle collisions were observed in any experiment.

Across the proposed-method trials, all trials that started disconnected reached $\lambda_2\geq\epsilon$, with initial maximum pairwise separations spanning $D_{\max}/R\in[2.31,5.84]$. No recovery failure was observed within this tested range.

Comparisons between our \emph{MPC--CLF--CBF} and \emph{CLF--CBF} highlight the benefit of planning, particularly in avoiding deadlocks in the presence of obstacles. Both methods maintain high connectivity without obstacles.
As obstacle density increases, \emph{MPC--CLF--CBF} generally achieves a higher success rate because the reactive \emph{CLF--CBF} controller encounters more frequent deadlocks and does not plan detours around obstacles.
\emph{MPC--CLF--CBF} also demonstrates higher connectivity in many obstacle-rich instances, as our approach plans over a finite horizon, thereby reducing infeasibility in the optimization problem. %

Compared to \emph{MPC--CBF}, our \emph{MPC--CLF--CBF} generally achieves a higher Percentage Connected in obstacle-rich environments, while success rate and makespan vary across settings. Specifically, without obstacles, both planners achieve high success rates and Percentage Connected scores with similar makespans. As obstacle density increases, our planner shows consistent improvements in Percentage Connected compared with the \emph{MPC--CBF} baseline, although it may incur a makespan trade-off as it prioritizes constraint satisfaction.
Overall, our \emph{MPC--CLF--CBF} planner demonstrates generally higher success rates and connectivity preservation rates while maintaining a comparable makespan.

Figure~\ref{fig:obstacle_case} illustrates a $10$-robot instance generated by our \emph{MPC--CLF--CBF} planner in an obstacle-rich workspace. 
Starting from an initially disconnected configuration, the team recovers connectivity by $t=\SI{5}{s}$ and later temporarily separates into two subgroups while detouring around obstacles at $t=\SI{10}{s}$--$\SI{12.5}{s}$.
The HOCLF recovery constraints subsequently bring the subgroups back within range; by $t=\SI{35}{s}$, connectivity is restored with $\lambda_2\geq\epsilon$, which remains above the threshold until all robots reach their goals.

\section{PHYSICAL EXPERIMENTS}
We also validate our algorithm with a team of $8$ Crazyflie nano-quadrotors inside a $10 \times 6$\,\si{m} workspace with a Vicon motion-tracking system. We fix the robots' heights and yaw angles. In the experiment, the robots are tasked with navigating through a cluttered environment. Since the Crazyflie quadrotors have limited onboard sensing and computational capabilities, we perform computation on a centralized computer and broadcast the control inputs through Wi-Fi to each quadrotor in real time. Non-circular obstacles in the physical experiments are conservatively approximated by enclosing circular regions to ensure safety. We provide the exact obstacle positions in our experiments, but in practice, they can be estimated through onboard perception, for example, using LiDAR or RGB-D cameras together with SLAM- or VIO-based state-estimation algorithms.

Figure~\ref{fig:demo} shows the executed trajectories of the quadrotors, demonstrating that the team maintains connectivity when feasible and achieves its navigation goal.

\section{CONCLUSION}

We present an HOCBF- and HOCLF-constrained motion-planning algorithm, namely MPC--CLF--CBF, that balances connectivity and task progress in obstacle-rich environments. Our algorithm generates B\'ezier-curve trajectories and the corresponding control inputs concurrently, making high-order derivatives directly available.  %
We validate the efficacy of our algorithm by benchmarking it against CLF--CBF and MPC--CBF baselines in various simulation instances, demonstrating improvements in task success rate and connectivity preservation. We further validate our motion planner through a physical demonstration with $8$ nano-quadrotors in a laboratory environment with obstacles.
Current work assumes global state access; future work will investigate computational scalability to larger teams and incomplete state information.
Another direction is to validate the framework in $3$D or move beyond double-integrator dynamics and incorporate full-body models for different robot types, paving the way for more realistic deployments of heterogeneous teams.

\balance
\bibliographystyle{IEEEtran}
\bibliography{refs} %

\end{document}